\documentclass{amsart}
\usepackage{amssymb,latexsym}

\theoremstyle{plain}

\theoremstyle{definition}

\theoremstyle{remark}

\numberwithin{equation}{section}

\begin{document}
\title[logic with verbs]{logic with verbs}
\author{Jun Tanaka}
\address{University of California, Riverside, USA}
\email{juntanaka@math.ucr.edu, junextension@hotmail.com}

\date{Dec, 14, 2009}

\begin{abstract}
The aim of this paper is to introduce a logic in which nouns and verbs are handled together as a deductive reasoning, and also to observe the relationship between nouns and verbs as well as between logics and conversations.
\end{abstract}

\maketitle


\section{\textbf{Introduction}}\label{SyloS:1}

In this paper, we will introduce a generalized logic which is capable
of handling nouns and verbs in statements. Statements in Classical Logic consists of two premises and one conclusion. For example,

Premise 1: X is a mother.

$\underline{\text{Premise 2: Every mother is a woman.    }}$

Conclusion: X is a woman.

The above structure of logic works mainly, with only primary verbs such as be and have. \cite{ling} In other word, Classical Logic is a logic for nouns. In
section 2, we will introduce a logic for nouns and verbs. We will
discuss how to widen the set of possible verbs, especially to lexical verbs, \cite{ling} and the structure of the presented modern logic is as follows:

Premise 1: a conditional statement for nouns

Premise 2: a conditional statement for verbs

$\underline{\text{Premise 3: a fact related to the above statements  }}$

Conclusion:

An example of the usage of this modern logic is as follows:

Premise 1: Tokyo is a part of Japan and Los Angeles is a part of U.S.

Premise 2: Flying is a way of traveling.

$\underline{\text{Premise 3: I flew from Tokyo to Los Angeles.  }}$

Conclusion: I traveled from Japan to U.S.

In Section \ref{SyloS:2}, we will introduce the main idea of this paper, Logic with Verbs. In Section \ref{SyloS:3} we will show its symbolic structure. In section \ref{SyloS:4}, we will observe the relationship between verbs and nouns.  Further, we will discuss how we mutually define verbs and nouns,
and also will present a potential application of modern logic to Fuzzy
Set Theory. In section \ref{SyloS:5}, we will introduce a method to relate subjects to verbs and nouns. In Section \ref{SyloS:6}, \ref{SyloS:7}, and \ref{SyloS:8}, we will discuss a way to apply this idea of
modern logic to studies in AI communication. Recently, a computability
of Natural Language is required especially in AI communication
theories. We will introduce one possible approach on this paper, which
we hope will be a productive contribution to AI in the future.

\section{\textbf{Logic With Verbs}}\label{SyloS:2}

\subsection{\textbf{The general form of Logic With Verbs}}\label{SyloS:genform}

In this section, we will relate nouns and verbs from a Set Theoretic view point. Please consider the following three chains;
\begin{itemize}
\item \textit{Orange $<$ fruit $<$ food $($Noun$)$}
\item \textit{Carrot $<$ vegetable $<$ food $($Noun$)$}
\item \textit{Fly $<$ Travel $<$ Move $($Verb$)$}
\end{itemize}

We will interpret the containments in Set Theory as specificities in order to generalize our usage. A Carrot is one kind of vegetable and vegetables are one kind of food. Similarly, to fly is one way to travel, to travel is one way to move. These are merely chains or orders of specificities, and this interpretation of specificity would be more suitable when we apply this Set Theoretic idea to a deductive reasoning as follows.

\begin{itemize}
 \item \textit{I flew from Tokyo to Los Angeles}
\item $\Rightarrow$ \textit{I traveled from Tokyo to Los Angeles} \\  \textit{(By considering Flying as a way of traveling)}
\item $\Rightarrow$ \textit{I traveled from Japan to U.S.} \\ \textit{ (By considering Tokyo $<$ Japan and Los Angeles $<$ U.S.)}
\end{itemize}

  Please note that the degree of meaning between the verbs fly, drive, run, and walk will depend on the relative distance to travel from Point A to Point B. Considering the above detailed example, flying is the most suitable way of traveling. This pattern of logic is applicable to the following verbs.

\begin{itemize}
\item \textit{Fly $<$ Travel}
\item \textit{drive $<$ Travel}
\item \textit{Walk $<$ Travel}
\item \textit{Run $<$ Travel}
\end{itemize}

We will give some examples of modern logic that is presented in this paper;

Example 1:

Premise 1: My brother is a lawyer.

Premise 2: Punching is a way of hitting

$\underline{\text{Premise 3: I punched my brother.   }}$

Conclusion: I hit a lawyer.

Example 2:

Premise 1: A sofa is furniture.

Premise 2: Wiping with a duster is a way of cleaning.

$\underline{\text{Premise 3: I wiped a sofa with a duster.   }}$

Conclusion: I cleaned furniture.

Example 3:

Premise 1: A potato is a vegetable.

Premise 2: Baking is a way of cooking.

$\underline{\text{Premise 3: I baked a potato.    }}$

Conclusion: I cooked a vegetable.

The author used the past tensed statement for Premise 3 in the above examples since normally facts are needed to be in the past. Thus, the past tensed statement should be used for Premise 3 in order to make the statement sound. However, please note that this structure works even with future or present tensed Premise statements as follows:

Example 4:

Premise 1: A sofa is furniture.

Premise 2: Wiping with a duster is a way of cleaning.

$\underline{\text{Premise 3: I will wipe a sofa with a duster.   }}$

Conclusion: I will clean furniture.

\subsection{\textbf{The negation in Logic With Verbs}}

In this section, we will introduce how to use negation in Logic With Verbs; The negation works basically the same as the classical logic.
\[
\text{Baking is a way of cooking} \Leftrightarrow \text{ Not cooking is a way of Not Baking}
\]

Thus, we have the following deductive reasoning:

Premise 1: A potato is a vegetable.

Premise 2: Baking is a way of cooking.

$\underline{\text{Premise 3:  I did not cook a vegetable.   }}$

Conclusion: I did not bake a potato.

\subsection{\textbf{Symbolic Structure of Logic With Verbs}}\label{SyloS:3}
The examples in the previous section of \ref{SyloS:genform} can be expressed in the following way:
For a fixed subject, let A,B, be verbs where A $<$ B and $<$ is a specificity. Let E,F be nouns where  and E $<$ F and $<$ is a specificity. Define a binary operator * by * : verb $\times$ noun $\longrightarrow$ a statement. So the binary operation create a collection of statements out of a collection of verbs and nouns. Then
\[
A* E  \Longrightarrow B*E \Longrightarrow B*F \ \text{or} \
A* E  \Longrightarrow A*F \Longrightarrow B*F
\]

Define a negation operator $\neg$ from a statement to a statement as follows:
\[
\neg B*F  \Longrightarrow \neg B*E \Longrightarrow \neg A* E \ \text{or} \
\neg B*F  \Longrightarrow \neg A*F \Longrightarrow \neg A* E
\]

In section \ref{SyloS:5}, we will discuss how to deal with different subjects. In this section, a subject simply must be attached in front of each statement. We will give an example when the subject is "you":
\[
\text{you} \ A* E  \Longrightarrow \text{you} \ B*E \Longrightarrow \text{you} \ B*F
\]

\subsection{\textbf{Symbolic Structure of And, Or}}\label{SyloS:andor}
When we say "I cooked vegetable and fruit", we are not thinking of an intersection of vegetable and fruits as in Classical Logic. We understand and define the statement "I cooked vegetable and fruit" as "I cooked vegetable and I cooked fruit". We will create the symbolic structure of And, Or.

For a fixed subject I,

Example 3 in the previous section can also be extended as in the form of right and left distributive laws as follows:

Let A,B,C,D be verbs where A $<$ B and C $<$ D and $<$ is a specificity. Let E,F,G,H be nouns where  and E $<$ F and G $<$ H and $<$ is a specificity. Then And and Or are defined between nouns and statements as well as between verbs and statements as follows:
\begin{itemize}
\item \text{A* (E and G) = A*E and A*G (Left distributive) For example,} \\
\text{I baked potatoes and apples = I baked potatoes and I baked apples.}
\item \text{(A and C)*E = A*E and C*E (right distributive) For example,} \\
\text{I baked and ate potatoes = I baked potatoes and I ate potatoes.}
\item \text{A* (E or G) = A*E or A*G (Left distributive) For example,} \\
\text{I baked potatoes or apples = I baked potatoes or I baked apples.}
\item \text{(A or C)*E = A*E or C*E (right distributive) For example,} \\
\text{I baked or ate potatoes = I baked potatoes or I ate potatoes.}
\end{itemize}

We can have a following application:
A* (E and G) = A*E and A*G $\Rightarrow$  B*E and B*G $\Rightarrow$ B*E and B*G

For example, I baked potatoes and apples = I baked potatoes and I baked apples $\Rightarrow$ I cooked vegetable and I cooked fruit = I cooked vegetable and fruit.

\section{\textbf{Mutual Definition of Nouns and Verbs}}\label{SyloS:4}

There are some pairs of verbs and nouns which are defined as a pair; that is where the relationship between nouns and verbs occurs. We call this N-V isomorphism. In this section, we will show how nouns and verbs should be related through a fuzzy set theoretic view. Some examples of N-V isomorphism as follow;

(1) Food is something you eat. Something you eat is most likely food. 

(2) A Beverage is something you drink. Something you drink is most likely a beverage.

(3) Something you ride on is a vehicle. A vehicle is something you ride on.

(4) Something you draw is a drawing. A drawing is something you draw.

(5) Something you sing is a song. A song is something you sing.

Eat and food are N-V isomorphic, and bread is food. Thus I can eat bread, and the statement "I can eat bread" is sound, (showing possibility). Now, we will show that N-V isomorphism is used to show the degree of possibility with fuzzy sets; Seaweed is food but if "I" is American, Seaweed is not very familiar as food. Thus the characteristic value of Seaweed as food must be low. Let's say 0.1. Then the statement "I can eat Seaweed" should be sound, but the statement "I rarely eat Seaweed" or "I am less likely to eat Seaweed" are more appropriate. Now some connections between N-V isomorphism and fuzzy sets are apparent.

So let's suppose the characteristic value of chicken as food is 0.95. "I often eat chicken" must be appropriate. We could let the range of characteristic values between 1-0.7 be "often", 0.7-0.4 be "more or less", 0.4-0.2 be "less likely", 0.2-0.05 be "rarely", 0.05-0 "never". Next we can create a Fuzzy Set Theoretic statement such as "I often eat pizza", "I rarely eat deer meat", and "I never eat a book" by following the method of Zadeh. \cite{Zad,Zad4}. 

\section{\textbf{Conditional Logic; How to deal with subjects}}\label{SyloS:5}
In this section, the author will present one possibility to handle subjects. Subjects make sentences subjective. In this interpretation, subjects affect and control the degree of possibility for doing X. In the previous section, the author mentioned "I rarely eat Sea Weed" or "I am less likely to eat Seaweed" if "I" is American. If "I" is Japanese, "I sometimes eat Sea Weed" or "I often eat Seaweed" must be appropriate. Thus, depending on the subject, the degree of possibility must vary.

\section{\textbf{Observation to Apply This Modern Logic to AI}}\label{SyloS:6}

In this section, we will compare daily conversations by using the presented modern logic. Our conversation never flows as Examples shown in Section 2. However, the author believes that the structure of the presented modern logic is necessary and applicable to AI communication. Natural Language does not need to give the most detailed information in our conversations, thus Natural Language provides only sufficient information or only a parts that he or she would like to emphasize. Then the listener may ask the speaker for more information if they are interested in more detail. I will give one example of a conversation which distinguishes flow of the presented modern logic.

Person A: "I traveled to U.S."

Person B: "Where in U.S. did you travel?"

Person A: "California"

Person B: "Where did you fly from?"

Person A: "I flew from Tokyo"

The above conversations sounds more natural than the examples presented in Section 2. Regular conversations typically go from a general statement to a more specific statement, depending on how much information is needed or how much interest is showed in, even while Logic flows from the specific statement to a more general statement. In order to make AI communicate ``humanistically'', we suggest generating the most specific statement for each fact beforehand, and then we must make it general enough to ``humanize conversations''. In other words, we need some filter on generated statements before the output of a statement.

\section{\textbf{An Application from the Observation of the Previous Section}}\label{SyloS:7}

Here is a systematized application for more natural conversations from the observation of Natural Language as  shown the previous section:

Premise 1: a house is a kind of a property

Premise 2: CA is a part of U.S.

Premise 3: buying X (for myself) is a way of owning X.

Premise 4: I will buy a house in CA. (a fact related to the above premises)

We will generate the below seven conclusions out of the four premises.

Conclusion 1: I will buy a house in U.S.

Conclusion 2: I will buy a property in CA.

Conclusion 3: I will buy a property in U.S.

Conclusion 4: I will own a house in CA.

Conclusion 5: I will own a house in U.S.

Conclusion 6: I will own a property in CA.

Conclusion 7: I will own a property in U.S.

In order to make this logic conversational, we need to reverse the pattern that is usually seen in the logic. We will demonstrate to generate a conversation between a computer program and a person and let Person A be a computer.

We call HOW, WHICH PART, WHAT KIND question operators which reverse A $<$ B. For example, if A $<$ B which means A is a kind of B, WHICH KIND * B = A, "WHICH KIND of property will you buy in CA?", the answer is "I will buy a house in CA." (WHICH KIND * property $\Rightarrow$ house.)

Person A "I will own property in U.S."

Person B "Which part of U.S. will you own property?"

Person A "I will own a property in CA" (WHICH PART * (own * property * U.S.) $\Rightarrow$ own * property * CA)

Person B "How will you own property in CA?"

Person A "I will buy property in CA" (HOW*(own*property*CA)$\Rightarrow$ buy*property*CA)

Person B "Which kind of property will you own in CA?"

Person A "I will buy a house in CA" (WHICH KIND*(buy*property*CA)$\Rightarrow$ buy*house*CA)

If Premises 1 to 4 are input beforehand in a program, it systematically generate correspondences just as above.

\section{\textbf{"If And Then" Statement In Logic With Verbs}}\label{SyloS:8}

We will introduce an extension of the application from the previous section, which shows how to handle "if and then" statements in Logic with Verb. In addition to premises 1 to 4 in the previous section, we will add one more premise as follows:

Premise 5: if I get this job, I will buy a house in CA.

then it implies all of the seven following conclusions.

Conclusion 1': If I get this job, I will buy a house in U.S.

Conclusion 2': If I get this job, I will buy a property in CA.

Conclusion 3': If I get this job, I will buy a property in U.S.

Conclusion 4': If I get this job, I will own a house in CA.

Conclusion 5': If I get this job, I will own a house in U.S.

Conclusion 6': If I get this job, I will own a property in CA.

Conclusion 7': If I get this job, I will own a property in U.S.

\section{\textbf{Conclusion and Observation}}\label{SyloS:conclusion}
In this paper, we tackled systematic expression of Linguistics especially for lexical verbs and nouns. This Modern Logic Theory would help us to bring systematic expression of languages closer to the level of sophistication of human conversations. I also strongly believe that this new logic system could open up a new branch of Artificial Intelligence. Further investigation in logic and linguistics are required to improve the systematic expression of our rational thought, which in turn is necessary in creating a communicative Artificial Intelligence. I dream of the day when we can create real AI.

\section{\textbf{Acknowledgement}}
The author would like to thank all professors who gave him very professional advice and suggestions, which he truly believes improved the preparation of this paper. The author gratefully acknowledges T. Hughes for his editorial assistance.




\begin{thebibliography}{99}

\bibitem{ling}
D. Biber, S. Johannson, G. Leech, S. Conrad, E. Finegan, Longman Grammar of Spoken and Written English, Longman Harlow, Essex ,UK 1999


\bibitem{Birk}
G. Birkhoff, Lattice Theory, 3rd ed. AMS colloquim Publication, Providence, RI, 1967.



\bibitem{being and time}
M. Heidegger, Being and Time, Harper One; Revised edition, 1962 (English)



\bibitem{Naka}
N. Nakajima, Generalized Fuzzy Sets, Fuzzy sets and Systems {\bf 32} (1989), 307-314.


\bibitem{Salmon}
W.C.Salmon, Logic, 3rd ed, Prentice Hall, 1984.

\bibitem{watanabe}
S. Watanabe, A Generalized Fuzzy Set Theory, IEEE Transactions on Systems, Man, and Cybernetics, vol SMC-8, No 10, p756, 1978.

\bibitem{witt}
L.Wittgenstein, Philosophical Investigations, 3rd ed, Prentice Hall, 1973.

\bibitem{Zad}
L.A. Zadeh, Fuzzy Sets, Information and Control {\bf8} (1965), 338-353.

\bibitem{Zad3}
L.A. Zadeh, Fuzzy Logic and its Application to Approximate Resoning, Information Processing {\bf74} (1974), pp 591-594.

\bibitem{Zad4}
L.A. Zadeh, The Concept of a Linguistic Variable and its Application to approximate Resoning II, Information Sciences {\bf8} (1975), 301-357.




\end{thebibliography}
\end{document}